\documentclass[lettersize,journal]{IEEEtran}
\usepackage{amsmath,amsfonts}
\usepackage{algorithmic}
\usepackage{array}
\usepackage[caption=false,font=normalsize,labelfont=sf,textfont=sf]{subfig}
\usepackage{textcomp}
\usepackage{stfloats}
\usepackage{url}
\usepackage{verbatim}
\usepackage{graphicx}

\usepackage{balance}

\IEEEoverridecommandlockouts    
\usepackage{cite}
\usepackage{amsmath,amssymb,amsfonts}
\usepackage{algorithmic}
\usepackage{setspace}
\usepackage{graphicx}
\usepackage{textcomp}
\usepackage{xcolor}
\usepackage{float}
\usepackage{url}

\usepackage{svg}
\usepackage{amsmath}

\usepackage{xcolor,colortbl}

\definecolor{aliceblue}{rgb}{0.94, 0.97, 1.0}

\usepackage{commath}
\usepackage{ulem}

\usepackage{hyperref}
\hypersetup{
    colorlinks=true,
    linkcolor=blue,
    filecolor=magenta,      
    urlcolor=cyan,
}

\graphicspath{ {.} }

\def\BibTeX{{\rm B\kern-.05em{\sc i\kern-.025em b}\kern-.08em
    T\kern-.1667em\lower.7ex\hbox{E}\kern-.125emX}}
\begin{document}


\title{S3TC: Spiking Separated Spatial and Temporal Convolutions with Unsupervised STDP-based Learning for Action Recognition\\}

\author{
\IEEEauthorblockN{
Mireille El-Assal\IEEEauthorrefmark{1},
Pierre Tirilly\IEEEauthorrefmark{1},
and Ioan Marius Bilasco\IEEEauthorrefmark{1}\\
}
\IEEEauthorblockA{\IEEEauthorrefmark{1}
\textit{Univ. Lille, CNRS, Centrale Lille, UMR 9189 CRIStAL, F-59000 Lille, France}\\
}
\IEEEauthorblockA{
Email: mireille.elassal2@univ-lille.fr, pierre.tirilly@univ-lille.fr, marius.bilasco@univ-lille.fr
}
}
\maketitle
\thispagestyle{plain}
\pagestyle{plain}

\begin{abstract}

Video analysis is a major computer vision task that has received a lot of attention in recent years. The current state-of-the-art performance for video analysis is achieved with Deep Neural Networks (DNNs) that have high computational costs and need large amounts of labeled data for training. Spiking Neural Networks (SNNs) have significantly lower computational costs (thousands of times) than regular non-spiking networks when implemented on neuromorphic hardware~\cite{Solving_spike_vanishing,TAVANAEI201947}. They have been used for video analysis with methods like 3D Convolutional Spiking Neural Networks (3D CSNNs). However, these networks have a significantly larger number of parameters compared with spiking 2D CSNN. This, not only increases the computational costs, but also makes these networks more difficult to implement with neuromorphic hardware. In this work, we use CSNNs trained in an unsupervised manner with the Spike Timing-Dependent Plasticity (STDP) rule, and we introduce, for the first time, Spiking Separated Spatial and Temporal Convolutions (S3TCs) for the sake of reducing the number of parameters required for video analysis. This unsupervised learning has the advantage of not needing large amounts of labeled data for training. Factorizing a single spatio-temporal spiking convolution into a spatial and a temporal spiking convolution decreases the number of parameters of the network. We test our network with the KTH, Weizmann, and IXMAS datasets, and we show that S3TCs successfully extract spatio-temporal information from videos, while increasing the output spiking activity, and outperforming spiking 3D convolutions.

\end{abstract}

\begin{IEEEkeywords}
spiking neural networks, STDP, action classification, 3D convolution, spatial, temporal, separated convolutions.
\end{IEEEkeywords}

\section{Introduction}
A large amount of new visual data is made available to the world on a daily basis, with a substantial portion of this data comprising videos. Analyzing this large amount of data is challenging for humans, which has rendered video analysis an important computer vision task. Deep learning methods achieve state-of-the-art performance for visual data analysis. However, their computational cost is very high, which makes using them on energy-constrained devices very challenging. Moreover, training them needs large amounts of labeled data, which requires costly human intervention to create. This has pushed forward the exploration of methods that can analyze visual data at a lower cost. Among these methods are Spiking Neural Networks (SNNs), which are third generation neural networks that can process visual information in the form of low-energy spikes~\cite{TAVANAEI201947}. They carry potential benefits such as fast information processing when implemented on neuromorphic hardware and energy efficiency~\cite{TAVANAEI201947,multLyrSNNWithTargetTmStampTrshAdpt, deepCSNN}, which encourages their use for video analysis.

There are some spiking models proposed for video analysis, including spiking two-stream methods~\cite{two-strRSNN}, spiking ResNets~\cite{NEURIPS2021_spikingResnets}, 2D Convolutional Spiking Neural Network (CSNN) and 3D CSNN~\cite{elassal:hal-03679597IJCNN}. While most spiking methods, similarly to 2D CSNNs, require non-spiking processes~\cite{elassal:hal-03263914} for motion extraction, 3D CSNNs~\cite{elassal:hal-03679597IJCNN} have the advantage of being fully spiking solutions for learning motion patterns. However, similarly to traditional methods, spiking 3D convolutions increase the number of trainable parameters with respect to spiking 2D convolutions. 
This can make it more challenging to implement this model on neuromorphic hardware, as having more parameters results in a greater number of connections that need to be constructed in the hardware implementation.
Consequently, this increases the computational costs of these networks as compared to 2D CSNNs. Therefore, there is a need to develop methods that can reduce the number of parameters while conserving the ability to directly learn spatio-temporal patterns. Separated convolution could be one solution, that has been used with CNNs~\cite{XSepConv,MobileNets,MobileNetV2}, but has not yet been used in the spiking domain. 

In this work, we present Spiking Separated Spatial and Temporal Convolutions (S3TCs), where we reduce the number of parameters in a spiking spatio-temporal 3D convolution by factorizing it into two separate smaller spatial and temporal convolutions. We use CSNNs trained with the unsupervised Spike Timing-Dependent Plasticity (STDP) learning rule. S3TCs are expected to be more efficient and hardware friendlier solutions. To the best of our knowledge, our work is the first to address the subject of separated convolutions with spiking neural networks. We hypothesize that the benefits of separated convolutions with CNNs could apply to SNNs, and that simpler filters, or smaller filter sizes, can improve STDP-based learning by capturing more patterns and prompting the neurons to fire more spikes. As STDP updates occur upon neurons firing spikes, this increased activity can result in more updates, making learning more effective.

This work is a building block towards improving the performance of spiking models that can learn spatio-temporal features. The main contributions of this paper are summarized as follows: 
\begin{itemize} 
  \item we present Spiking Separated Spatial and Temporal Convolutions (S3TCs);
  \item we evaluate the performance of S3TC models with different filter sizes on the \href{https://www.csc.kth.se/cvap/actions/}{KTH}~\cite{b39}, \href{http://www.wisdom.weizmann.ac.il/~vision/SpaceTimeActions.html}{Weizmann}~\cite{b37}, and \href{https://www.epfl.ch/labs/cvlab/data/data-ixmas10/}{IXMAS}~\cite{weinland:inria-ixmas} datasets;
  \item we compare the performance of S3TCs to that of spiking 3D convolution from \cite{elassal:hal-03679597IJCNN}, and we conclude that S3TCs can achieve better performance;
  \item we show that smaller filter sizes, to a certain extent with STDP, prompt the neurons to fire more spikes, thus increasing the network activity and improving the learning. 
\end{itemize}

\section{Related Work}
\label{section:literature}
 
3D CNNs are a common practice for motion modeling~\cite{Romain3DWACV,baccouche:hal-01354493,can3Dret2D,slowfast_ICCV,X3D,kth3d11frames}. The third dimension of these networks, which is devoted to time, enables the extraction of spatio-temporal features. In~\cite{LearningSPTFeatures}, the authors present deep 3D CNNs for spatio-temporal feature learning. They compare them to 2D CNNs, and conclude that 3D architectures perform better for video analysis.  However, these models still have many issues. For instance, they have more trainable parameters than 2D models, which consequently increases their computational cost and makes the optimization of these parameters more difficult. Moreover, these models are energy-consuming and require powerful GPUs to run efficiently. Running them on devices with limited energy is very challenging. The environmental concerns of running powerful GPUs for extended periods of time~\cite{AI_carbon, strubell2019energy} motivated the search for alternative methods that have lower computational costs, one of which is separable convolutions. With separable convolutions, a large convolution filter is separated into two or more smaller filters. Separable convolutions adopted in networks like MobileNets~\cite{MobileNets} and Xception~\cite{Xception} have succeeded in decreasing the number of parameters of these networks while preserving their performance. Moreover, gains in accuracy have been recorded when factorizing a 3D convolution into a 2D spatial convolution and a 1D temporal convolution~\cite{tran2018closer}. In~\cite{tran2018closer}, the authors attribute this gain in accuracy to additional nonlinearities added by the separated convolutions compared to using a 3D convolution. They argue that these nonlinearities render the model capable of representing more complex functions. They also add that 2D and 1D filters are easier to optimize than 3D filters, where appearance and dynamics are intertwined. However, none of the neural networks discussed up until this point are spiking models. 

CSNNs provide a cost-effective and unsupervised alternative for motion modeling. 3D CSNNs have been proposed recently~\cite{elassal:hal-03679597IJCNN}. In~\cite{elassal:hal-03679597IJCNN}, the authors use unsupervised STDP-based 3D CSNNs, and conclude that 3D CSNNs outperform 2D CSNNs for the task of human action recognition, especially with longer video sequences. However, despite the energy efficiency of these 3D CSNN models compared with traditional non-spiking methods, the additional parameters, with regard to 2D CSNNs, result in higher costs and potentially more complex neuromorphic hardware~\cite{loihi}. As previously mentioned, spiking-separable convolutions have not yet been explored in the literature. However, they are excellent candidates for efficient hardware-friendly video analysis models.

\section{Background \& Network Architecture}
\label{section:networkarch}

This section contains the needed background information and mechanisms chosen for achieving S3TCs with unsupervised STDP learning. This includes a simplified explanation of 3D CSNNs from \cite{elassal:hal-03679597IJCNN}, and a discussion of the number of parameters required to train this network.

\begin{figure*}
\centerline{\includegraphics[scale=0.28]{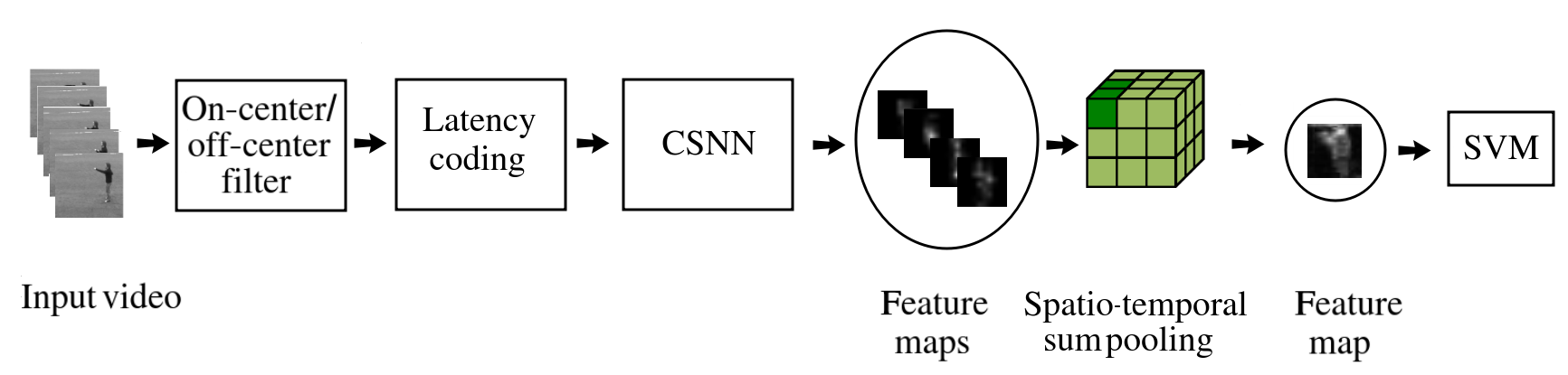}}
\caption{Pipeline for action recognition with unsupervised feature learning by an SNN.}
\label{figNetwork3D}
\end{figure*}

\subsection{A video sample}
\label{subsection:video}
A video is a frame sequence represented as a 4D tensor of size $l_{w} \times l_{h} \times l_{c} \times l_{td}$ where $l_{w}$ and $l_{h}$ are the width and height of the frames, $l_{c}$ is their number of channels, which is $1$ in the case of gray scale frames, and $l_{td}$ is the temporal depth of the tensor i.e., the number of frames in the video sample.

\subsection{Neuron model and training}
\label{subsection:neuronmodel}
The CSNNs used in this work consist of one feed-forward layer that contain Integrate-and-Fire (IF) neurons~\cite{IFneuron}. The IF neuron model is characterized by having a certain membrane potential $v(t)$ and a threshold potential $v_{\text{th}}(t)$. Input spikes are integrated into the neuron's membrane potential until it reaches its threshold, thereby triggering the neuron to generate an output spike. Then, the membrane potential is reset to its resting potential $v_{r}$, which is $0$ volts in this work. Specifically, we have built upon the network architecture of~\cite{elassal:hal-03679597IJCNN} and utilized the same neuron model and threshold adaptation mechanisms. This decision was made to ensure an accurate comparison of the performance between our S3TC and the 3D convolution model presented in~\cite{elassal:hal-03679597IJCNN}.

Training is unsupervised and uses the biological STDP learning rule~\cite{SchumanSTDP}. This is a local learning rule that acts on a synapse connecting two neurons at a time. Therefore, it is not affected by changes in the network architecture. These networks use Winner-Takes-All (WTA) inhibition during training to prevent several neurons from learning the same pattern. With WTA, some neurons can overpower other neurons, i.e., they have a tendency to fire more spikes than others. This leads to the network becoming stuck in a state where a few active neurons fire all the time, while the other neurons are quiet. In order to ensure the stability of the network, we use the threshold adaptation method introduced in~\cite{multLyrSNNWithTargetTmStampTrshAdpt}. This method trains the neurons to fire at a given target timestamp $\hat{t}$, which controls the patterns learned by the network, and promotes the activity of all neurons. Within this method, each time a neuron fires or receives an inhibitory spike, the thresholds of all neurons (winners and losers) are adapted so that their firing time converges rapidly towards $\hat{t}$. Furthermore, the threshold of the neuron that fired is raised, while the thresholds of all the other neurons are lowered, with the aim of encouraging their activation.

\subsection{3D convolution}
\label{section:3dconv}

A 3D convolutional layer has $f_{k}$ trainable filters, with sizes $f_{w} \times f_{h} \times f_{\text{td}}$, where $f_{w}$ and $ f_{h}$ represent the width and height of the filter respectively, and $f_{\text{td}}$ is the temporal size of the filter. These filters can slide along the temporal dimension of a video sample, in addition to the spatial ones.

Each neuron of a layer is connected to $f_{w} \times f_{h} \times f_{\text{td}}$ neurons of the previous layer. 3D spiking convolution can be formalized as shown in Equations~\ref{eq:filter}~and~\ref{eq:3DConv} from~\cite{elassal:hal-03679597IJCNN}: 

\begin{equation} \label{eq:filter}
f_{s}(x) = 
\begin{cases}
    1, & \mbox{if } x \geq 0 \\ 
    0, & \mbox{otherwise}
\end{cases}
\end{equation}

\begin{equation} 
\label{eq:3DConv}
v_{x,y,z,k}(t) =\sum_{n \in \mathcal{N}} W_{i(x_{n}), j(y_{n}), m(z_{n}), k_{n}, k} \times f_{s}(t-t_{n})
\end{equation} 

\noindent where $f_{s}$ is the kernel of spikes, $v(t)$ is the potential of the neuron membrane at time $t$,  and $x$, $y$, $z$, and $k$ are the coordinates of the spike in the width, height, time, and channel dimensions, respectively. $\mathcal{N}$ is the set of input connections in the neighborhood, $W \sim U(0, 1)$ is the trainable synaptic weight matrix, $i()$, $j()$, and $m()$ are functions that are used to map the location of the input neuron to the corresponding location in the weight matrix, and $k$ is the index of the trainable filter. When the membrane potential $v_{x,y,z,k}(t)$ crosses the threshold potential $v_{\text{th}}(t)$, the synaptic weights and thresholds of the network are updated.

The number of parameters required for training a spiking 3D CSNN is:
\begin{equation} \label{eq:nbrparam3d}
|P| = f_{k} \times n_{\text{c}} \times f_{w} \times f_{h} \times f_{\text{td}}
\end{equation}

\noindent where P represents the set of parameters in the model, $f_{k}$ is the number of filters, $n_{\text{c}}$ is the number of input channels, $f_{w}$ and $ f_{h}$ represent the width and height of the filter, respectively, and $f_{\text{td}}$ is the temporal size of the filter.

\subsection{Baseline architecture}
\label{subsection:baselinearch}

The baseline architecture is shown in Figure~\ref{figNetwork3D}. The video frames are filtered with an on-center/off-center filter~\cite{OnOffCS}, which uses a Difference-of-Gaussian (DoG) filter used to pre-process the data by simulating on-center/off-center cells and extracting edges. This filter is needed because STDP-based SNNs need edges to learn informative patterns~\cite{ImprSNNTrain}. After that, latency coding is applied to transform the edges into spikes represented by timestamps. High edge values are represented as early spikes, while low values come later. Next comes the CSNN processing, which can be a 3D CSNN as mentioned in Section~\ref{section:3dconv} or a S3TC network. The output of our network will consist of spatio-temporal feature maps, which are then reduced in size using spatio-temporal sum-pooling before being sent to the SVM for classification.

\section{Separated spatial \& temporal convolutions}
\label{subsection:sepconv}

\begin{figure}[H]
\centerline{\includegraphics[scale=0.28]{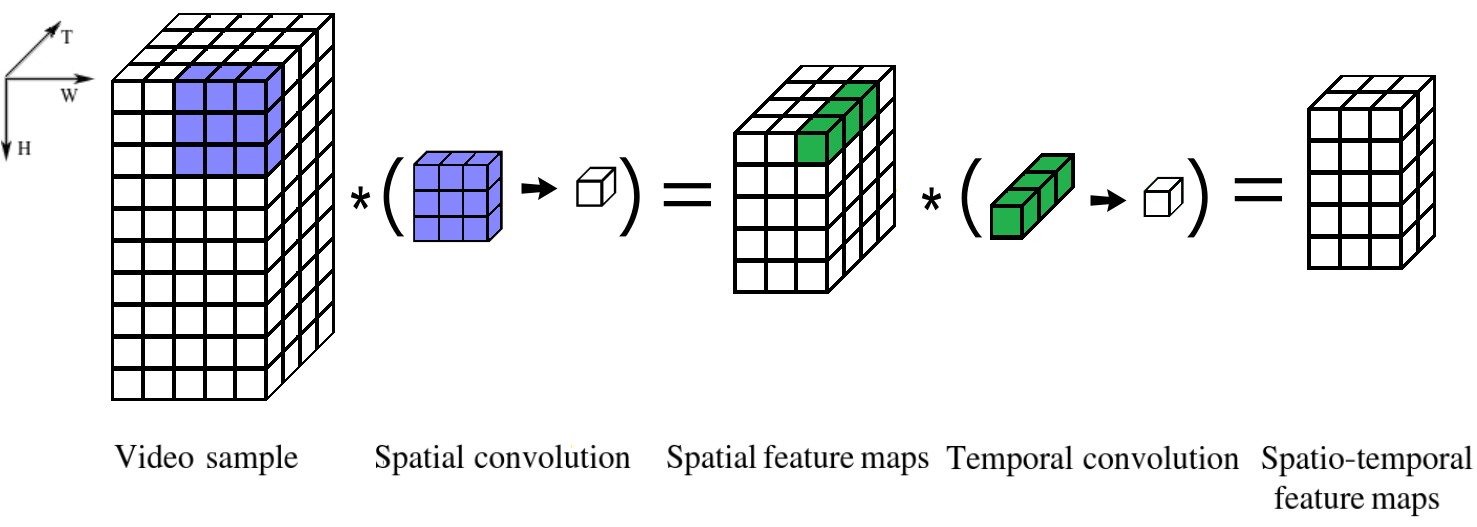}}
\caption{Separable spatial and temporal convolutions.}
\label{fig:2D1Dconv}
\end{figure}

With separated convolutions, the filter connectivity of the spiking 3D convolution layer introduced in Section~\ref{section:3dconv} can be broken down into two parts, space-wise and time-wise convolutions, as shown in Figure~\ref{fig:2D1Dconv}. 

In the first phase, a 2D filter crosses over the spatial dimension of the input, one frame at a time. This filter has a dimension of $f_{w} \times f_{h} \times 1$, and results in spatial feature maps. In the second phase, with the time-wise convolution, we compute a linear combination of the spatial feature maps by undergoing a $1 \times 1 \times f_{\text{td}}$ convolution in the temporal dimension to extract meaningful temporal information from the spatial feature maps. S3TC can be formalized as Equation~\ref{eq:spatialconv} for the space-wise convolution, and Equation~\ref{eq:tempconv} for the time-wise convolution:

\begin{equation}\label{eq:spatialconv}
v_{x,y,k}(t) =\sum_{n \in \mathcal{N}_{s}} W_{i(x_{n}), j(y_{n}), k_{n}, k} \times f_{s}(t-t_{n})
\end{equation}
\begin{equation}\label{eq:tempconv}
v_{z,k}(t) =\sum_{n \in \mathcal{N}_{t}} W_{m(z_{n}), k_{n}, k} \times f_{s}(t-t_{n})
\end{equation}

\noindent where $f_{s}$ is the kernel of spikes, $v(t)$ is the potential of the neuron membrane at time $t$,  and $x$, $y$, $z$, and  $k$ are the coordinates of the spike in the width, height, time, and channel dimensions, respectively. $W \sim U(0, 1)$ is the trainable synaptic weight matrix, $i()$, $j()$, and $m()$ are functions that are used to map the location of the input neuron to the corresponding location in the weight matrix, $k$ is the index of the trainable filter, and $\mathcal{N}_{s}$ and $\mathcal{N}_{t}$ are the sets of input connections in the spatial and temporal neighborhoods, respectively.

The number of parameters required for training S3TCs becomes:
\begin{equation} \label{eq:nbrparam2d1d}
|P'| = f_{k}  \times n_{\text{c}} \times (f_{w} \times f_{h} + f_{\text{td}})
\end{equation}

This number of parameters is lower than that of a spiking 3D convolution, therefore the computational cost of training S3TCs is lower than that of training spiking 3D convolutions. In the next section, we study the trade-off between accuracy and efficiency with these two spiking convolution settings.

\section{Evaluation}
\label{section:evaluation}
This section contains the details of our experiments. First, we present the datasets, along with the implementation details and the main parameters of our network. Then we present the results of implementing and testing our S3TCs, and we compare them to spiking 3D convolutions.

\subsection{Datasets and evaluation protocol}
\label{subsection:dataset}
We use three datasets, the KTH, Weizmann, and IXMAS datasets. The KTH and Weizmann datasets are early and simple datasets for action recognition. Although traditional computer vision approaches have already achieved high recognition rates on these datasets~\cite{KTHGoodPErcent}, their simplicity makes them good candidates to study the performance of new models like the ones targeted in this paper. The IXMAS dataset features different actors, cameras, and viewpoints, which adds complexity. Moreover, the settings are challenging, as two thirds of the recordings contain objects in the scene, partially occluding the actors.

The KTH dataset contains 600 videos that are made up of 25 subjects performing 6 actions in 4 scenarios. Subjects 11, 12, 13, 14, 15, 16, 17, and 18 are used for training, while 19, 20, 21, 23, 24, 25, 01, 04 are used for validation, and 02, 03, 05, 06, 07, 08, 09, 10, and 22 are used for testing, as indicated in the KTH protocol.

The Weizmann dataset contains 90 videos of 9 subjects performing 10 actions. The experiments on this dataset all use the leave-one-subject-out strategy. 

The IXMAS action recognition dataset is made up of 10 subjects, 11 actions and 1148 sequences. The experiments on this dataset also use the leave-one-subject-out strategy. 

To shorten the running time of experiments, we take subsets of the video frames, like in~\cite{kth3d11frames},~\cite{3DCNNHAR}, and~\cite{elassal:hal-03679597IJCNN}. We use $10$ frames per video, and skip three frames between each two selected frames in order to make sure to capture a full cycle of the performed action. We also scale down the frame sizes to half of their original sizes to increase the processing speed.

We measure the classification accuracy (in \%) on the test set for all experiments. Each experiment was run three times, and we report the average accuracy over the three runs.

\subsection{Implementation details}
\label{subsection:impdetails}
The video is pre-processed with the on-center/off-center filter mentioned in~\ref{subsection:baselinearch}. This filter has a size of $7$, and uses centered Gaussians of variance $1.0$ and $4.0$. The convolutional layer has $f_{k} = 64$ filters for both 3D and S3TC settings.

The neuron thresholds are randomly initialized with a normal distribution, which has a mean of $8$ and variance of $0.1$ for all experiments except those with a filter size of $3$, where we decrease the mean to $7$. This is because small filters integrate fewer input spikes, resulting in no spiking activity when the threshold is too high. 

The value of the target timestamp $\hat{t}$ discussed in Section~\ref{subsection:neuronmodel} are taken from~\cite{elassal:hal-03679597IJCNN}. We use a value of $\hat{t} = 0.65$ for the KTH and IXMAS datasets, and a value of $\hat{t} = 0.75$ for the Weizmann dataset. 
The spatio-temporal pooling is set to limit the size of the output feature maps to $20 \times 20 \times 2$. 

Then the output feature maps are linearized and introduced into a Support Vector Machine (SVM) with a linear kernel, which performs action classification. Any other supervised method can be used for the final classification; we chose an SVM because it is standard and effective with default hyperparameters.

The software simulator used to simulate the convolutional SNNs tested in this work is the \href{https://gitlab.univ-lille.fr/bioinsp/falez-csnn-simulator/tree/07fd14324afc42d7b3b24a3472271e1c6a90255a}{csnn-simulator}~\cite{ImprSNNTrain}, which is a publicly available and open-source simulator. The source code for our experiments will be released publicly as a specific branch of the csnn-simulator~\cite{ImprSNNTrain}.

\subsection{3D vs. Separable convolutions}\label{sepConv}
We test 3D convolutions and S3TCs for five different filter sizes. For the sake of limiting the possible filter size combinations, we use the same size $f = f_{h} = f_{w} = f_{td}$ for all dimensions. A 3D convolution has filters of size $f \times f \times f$, while the filter sizes of its corresponding separated convolutions are $f \times f \times 1 $ for the spatial convolution and $1 \times 1 \times f$ for the temporal one. The most commonly used filter size is $3$~\cite{Xception, MobileNets, MobileNetV2}. However, larger filter sizes like $5$ and $7$ have shown to give better results in~\cite{MobileNetV3}, so we include these filter sizes in our experiments.

\begin{table}
\begin{center}
\begin{tabular}{|c | c | c |}  
\hline
\rowcolor{aliceblue} \multicolumn{3}{|c|}{(A) Filter size = 3}\\ 
\hline
\rowcolor{aliceblue}Dataset  & 3D Conv & Separated Conv  \\
\hline
KTH &        59.41 $\pm$ 0.43 & 60.65  $\pm$  0.37 \\ [0.5ex] 
\hline
Weizmann &   56.58 $\pm$  0.00 & 60.06  $\pm$  1.88 \\ [0.5ex] 
\hline
IXMAS &      53.81 $\pm$ 0.45 & 52.26  $\pm$  0.63 \\ [0.5ex] 
\hline

\rowcolor{aliceblue} \multicolumn{3}{|c|}{ (B) Filter size = 5 }\\ 
\hline
\rowcolor{aliceblue}Dataset  & 3D Conv & Separated Conv \\
\hline
KTH &        61.88 $\pm$ 0.57 & 69.29  $\pm$  0.21\\ [0.5ex] 
\hline
Weizmann &  57.61 $\pm$ 0.00 & 66.24  $\pm$  0.00 \\ [0.5ex] 
\hline
IXMAS &     51.56 $\pm$ 0.12 & 51.44  $\pm$  0.18  \\ [0.5ex] 
\hline

\rowcolor{aliceblue} \multicolumn{3}{|c|}{ (C) Filter size = 7 }\\ 
\hline
\rowcolor{aliceblue}Dataset  & 3D Conv & Separated Conv \\
\hline
KTH &        64.20 $\pm$ 0.21 & 70.52  $\pm$  0.78 \\ [0.5ex] 
\hline
Weizmann &   57.09 $\pm$ 0.00 & 65.78  $\pm$  0.92 \\ [0.5ex] 
\hline
IXMAS &      40.74 $\pm$ 0.38 & 48.87  $\pm$  0.60 \\ [0.5ex] 
\hline

\rowcolor{aliceblue} \multicolumn{3}{|c|}{ (D) Filter size = 9 }\\ 
\hline
\rowcolor{aliceblue}Dataset  & 3D Conv & Separated Conv \\
\hline
KTH &        62.81  $\pm$  0.21 & 65.43  $\pm$  0.21 \\ [0.5ex] 
\hline
Weizmann &   56.50  $\pm$ 0.00 & 64.62  $\pm$  0.00  \\ [0.5ex] 
\hline
IXMAS &      34.01  $\pm$  0.27 & 46.46  $\pm$  0.21 \\ [0.5ex] 
\hline

\rowcolor{aliceblue} \multicolumn{3}{|c|}{ (E) Filter size = 10 }\\ 
\hline
\rowcolor{aliceblue}Dataset  & 3D Conv & Separated Conv \\
\hline
KTH &        60.65 $\pm$  0.00 & 61.11  $\pm$  0.75  \\ [0.5ex] 
\hline
Weizmann &   58.09 $\pm$ 0.52 & 57.12  $\pm$  0.52\\ [0.5ex] 
\hline
IXMAS &      27.94 $\pm$ 0.20 & 38.50  $\pm$  0.28 \\ [0.5ex] 
\hline
\end{tabular}
\end{center}
\caption{Classification rates in \% (average $\pm$ standard deviation) for the KTH, Weizmann, and IXMAS datasets (10 frames per video) over 3 runs with 3D convolution and separated convolutions.}
\label{table:3DvsSep}
\end{table}

Table~\ref{table:3DvsSep} shows the results of the experiments. These results show that S3TCs can achieve better performance than 3D convolution, while having less parameters, and thus a lower computational cost. These results show that different datasets have different ideal filter sizes for S3TCs ($7$ for the KTH dataset, $3$ for IXMAS, and $5$ for Weizmann). These results also show that filter size has more impact on the results with S3TC; for instance, with the KTH dataset, changing the filter size from $3$ to $7$ using 3D convolutions has an impact of approximately $+5 p.p.$, while with S3TCs, the impact is approximately $+10 p.p$.

In the context of action recognition, the spatial filters are responsible for extracting interesting spatial features in successive frames. However, these filters alone cannot make sense of the relationship between these extracted feature maps. 
The 1D filters model the time dimension, thus combining the output feature maps of the 2D filters into more discriminant combinations that include motion. Larger temporal filters provide a better extraction of these moving patterns with datasets that exhibit significant variations or movements relative to the frame size, like the KTH dataset, while smaller filters are needed for datasets that exhibit smaller variations, like the Weizmann dataset. Therefore, the performance of S3TCs, similarly to 3D convolutions, depends greatly on choosing suitable hyperparameters.

With large enough filter sizes (e.g., $7$ and $9$), the S3TC method can outperform regular 3D convolutions for most datasets. This behavior, where separated convolutions outperforms 3D convolutions, is similar to the one observed in~\cite{tran2018closer}, where they specify that 2D and 1D filters are easier to optimize with supervised learning than 3D filters. 

\begin{figure}
\centerline{\includegraphics[scale=0.5]{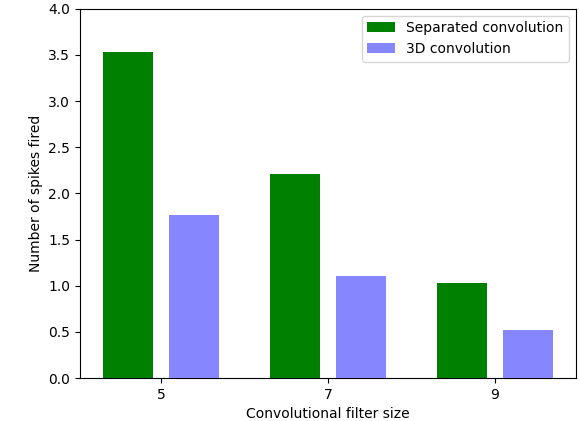}}
\caption{The final output number of spikes fired by separable spatial and temporal convolutions compared to 3D convolutions with filter sizes of $5$, $7$ and $9$ using the KTH dataset.}
\label{fig:spikesfired}
\end{figure}

In our case, this supervised weight optimization process does not happen with STDP, where we use unsupervised learning, and thus cannot be the reason for the amelioration in performance. In the case of a spiking model, we observe that more spikes are fired when using separated convolutions than its 3D counterpart, as shown in Figure~\ref{fig:spikesfired}. 
This is due to the simplicity of the 2D spatial and 1D temporal filters, which enables them to capture more pattern occurrences than the complex 3D filters. This results in firing a larger number of output spikes with S3TCs and increases the activity of the network. This yields less sparsity in the final vector fed into the SVM and more training updates with STDP, which results in a higher accuracy. This is consistent with previous observations that excessive sparsity can lower recognition performance~\cite{b3}.


\section{Conclusion}
\label{section:conclusion}
Spiking neural networks can offer an energy-efficient solution on neuromorphic hardware. However, using 3D convolutions, which are suitable for video analysis, increases the number of parameters, making training more challenging and potentially leading to more complex hardware requirements. To mitigate this issue, we chose to reduce the number of parameters in the network by replacing spiking 3D convolutions with spiking separated convolutions. 

In this work, we factorize a single 3D spiking convolution into two separate spatial and temporal spiking convolutions. This separation decreases the number of parameters, and can improve the performance when using sufficiently large filters. The difference in performance between 3D convolutions and separable convolutions is highly dependent on choosing the appropriate hyperparameters (i.e., filter size). 

Our first conclusion is that the optimum filter sizes vary from one dataset to another depending on their motion variations. A second conclusion is that S3TCs can outperform 3D convolutions due to the simplicity of their filters, which leads to capturing more patterns and thus firing more spikes. We believe that, with STDP, there is a proportional relationship between the number of input spikes in a layer and the quality of the learned filters. Additionally, the fact that S3TCs produce more output spikes makes them more advantageous than 3D CSNNs for constructing multi-layer spiking models, since activity loss is a major issue in deep SNNs~\cite{MasteringOutputFrequency}.

A promising avenue for future work would involve using a multi-stream architecture with S3TC networks, each stream using a specific filter size. This approach would enable capturing information about both small and large motion patterns, resulting in better generalization across different datasets.

\section{Acknowledgments}
This work has been supported by IRCICA (USR 3380) under the bio-inspired project, and funded by Région Hauts-de-France. 

\bibliographystyle{unsrt}
\bibliography{CitationLibrary}
\vspace{12pt}
\color{red}
\end{document}